# Using arguments for making decisions:
# A possibilistic logic approach

**Leila Amgoud**
IRIT - UPS
118, route de Narbonne
31062, Toulouse, FRANCE
E-mail: amgoud@irit.fr

**Henri Prade**
IRIT - UPS
118, route de Narbonne
31062, Toulouse, FRANCE
E-mail: prade@irit.fr

## Abstract

Humans currently use arguments for explaining choices which are already made, or for evaluating potential choices. Each potential choice has usually pros and cons of various strengths. In spite of the usefulness of arguments in a decision making process, there have been few formal proposals handling this idea if we except works by Fox and Parsons and by Bonet and Geffner. In this paper we propose a possibilistic logic framework where arguments are built from an uncertain knowledge base and a set of prioritized goals. The proposed approach can compute two kinds of decisions by distinguishing between pessimistic and optimistic attitudes. When the available, maybe uncertain, knowledge is consistent, as well as the set of prioritized goals (which have to be fulfilled as far as possible), the method for evaluating decisions on the basis of arguments agrees with the possibility theory-based approach to decision-making under uncertainty. Taking advantage of its relation with formal approaches to defeasible argumentation, the proposed framework can be generalized in case of partially inconsistent knowledge, or goal bases.

## 1  Introduction

In everyday life, decision is often based on arguments and counter-arguments. The decisions made in this way have a basis that can be more easily referred to for explanation purposes. Such an approach has indeed some obvious benefits. On the one hand, a best choice is not only suggested to the user, but also the reasons of this recommendation can be provided in a format that is easy to grasp. On the other hand, such an approach to decision making would be more acute with the way humans often deliberate and finally make a choice. More generally, argumentation systems (e.g. [1, 2, 6, 14, 17, 20]) have been developed in AI and proved to be useful in a variety of tasks. In particular, argumentation is a promising model for reasoning with inconsistent knowledge. It follows a three steps process: constructing arguments and counter-arguments, then selecting the most acceptable of them, and finally concluding.

The idea of basing decisions on arguments pro and con was already advocated more than two hundreds years ago by Benjamin Franklin [18]. This idea has been also at work among critical thinking philosophers. However, there has been almost no attempt at formalizing it until now if we except some recent works by Fox and Parsons [17] (see [16] for an informal and introductory discussion) and by Bonet and Geffner [4]. However, these works suffer from some drawbacks: the first one being based on an empirical calculus while the second one, although more formal, does not refer to argumentative inference. So there is a need for a more general approach where inconsistency is handled in an argumentative logic manner and which agrees with a qualitative view of decision under uncertainty.

In order to keep the benefit of existing axiomatic justifications, the proposed approach is a counterpart, in terms of logical arguments, of the possibilistic qualitative decision setting (which has been axiomatized both in the von Neumann [8] and in the Savage styles [13]). Moreover, a logical representation of the possibilistic decision framework [7] has been developed, where both the available knowledge (which may be pervaded with uncertainty) and the goals representing the user preferences (with their priority levels) are encoded by two distinct possibilistic logic bases. From these logical bases, it should be possible to build the different arguments in favour and against a decision and to compute their strengths. This framework distinguishes between pessimistic and optimistic attitudes toward risk. This gives birth to different types of arguments in favour and against a possible choice.

The paper is organized in the following way. First, the possibilistic logic counterpart of possibility theory-based decision is recalled in section 2. Then Section 3 presents the de-cision procedure in terms of arguments-based evaluations both in the pessimistic and in the optimistic cases, when the knowledge base and the goal base are both consistent. Section 4 discusses possible extensions of the approach when these bases become inconsistent. Section 5 provides a comparison with Fox-Parsons and Bonet-Geffner approaches.

## 2 Logical handling of qualitative decision under uncertainty

In what follows, $\mathcal{L}$ denotes a propositional language, $\vdash$ classical inference, and $\equiv$ logical equivalence.

In decision under uncertainty, possibilistic logic can be used for modeling the available information about the world on the one hand, and the preferences on the other hand. This section summarizes a proposal made in [7, 12] for designing a logic-based decision machinery. We distinguish between two possibilistic logic bases. The first one $\mathcal{K} = \{(k_j, \rho_j); j = 1, l\}$ represents the available knowledge about the world. $k_j$ is a proposition of the language $\mathcal{L}$ and the pair $(k_j, \rho_j)$ is understood as $N(k_j) \geq \rho_j$, where $N$ is a necessity measure [10]. Namely $(k_j, \rho_j)$ encodes that the piece of knowledge "$k_j$ is true" holds as certain at least at level $\rho_j$, where $\rho_j$ belongs to a linearly ordered valuation scale $R$ where there exist a top and a bottom element which are respectively denoted by 1 and 0.

The second possibilistic logic base $\mathcal{G} = \{(g_i, \lambda_i); i = 1, m\}$ represents the preferences of the decision-maker under the form of a prioritized set of *goals*, where $g_i$ is a proposition of the language $\mathcal{L}$ and $\lambda_i$ is the level of priority for getting the goal $g_i$ satisfied. Priority levels take their values on another linearly ordered scale $T$ with top and bottom denoted $\overline{1}$ and $\overline{0}$.

We shall denote by $\mathcal{K}^*$ and $\mathcal{G}^*$ the corresponding sets of classical propositions when weights are ignored.

The propositional language contains Boolean decision variables and Boolean state variables. In this setting a decision is a conjunction of decision literals, denoted $d$. In this view a do-nothing decision is represented by a tautology. The set $\mathcal{D}$ gathers all the allowed decisions. Each potential decision $d$ is represented by a formula $(d, 1)$ to be added to $\mathcal{K}$ if the decision is chosen. Let $\mathcal{K}_d = \mathcal{K} \cup \{(d, 1)\}$ be the description of what is known about the world when $d$ is applied. Associated with the possibilistic logic base $\mathcal{K}_d$ is the possibility distribution $\pi_{\mathcal{K}_d}$ which defines its semantics counterpart

$$\pi_{\mathcal{K}_d}(\omega) = min_{j=1,l} max(v_\omega(k_j), n_R(\rho_j)),$$

which rank-orders the more or less plausible states of the world when $d$ is chosen, where $v_\omega(k_j) = 1$ if $\omega$ is a model of $k_j$ and $v_\omega(k_j) = 0$ if $\omega$ falsifies $k_j$ and where $n_R$ is the order-reversing map of $R$ (see [10] for details).

Associated with the layered set of goals $\mathcal{G}$ is the ordinal utility function

$$\mu_\mathcal{G}(\omega) = \min_{i=1,m} \max(v_\omega(g_i), n_T(\lambda_j)),$$

which rank-orders the different states according to their acceptability, and where $n_T$ is the order-reversing map of $T$. $\pi_{\mathcal{K}_d}$ and $\mu_\mathcal{G}$ are assumed to be normalized ($\exists \omega, \pi_{\mathcal{K}_d}(\omega) = 1; \exists \omega', \mu_\mathcal{G}(\omega') = 1$), which is equivalent to the consistency of $\mathcal{K}^*$ and $\mathcal{G}^*$.

From $\pi_{\mathcal{K}_d}$ and $\mu_\mathcal{G}$, a pessimistic qualitative utility (see Appendix) can be computed as

$$E_*(d) = \min_\omega \max(\mu_\mathcal{G}(\omega), n(\pi_{\mathcal{K}_d}(\omega))) \qquad (1)$$

where $n$ is a decreasing map from $R$ to $T$ such that $n(0) = \overline{1}$ and $n(1) = \overline{0}$. In the following we assume the full commensurateness of the scales (i.e. $R = T$ and $n_R = n_T = n$). $E_*(d)$ is all the greater as all the plausible states $\omega$ according to $\pi_{\mathcal{K}_d}$ are among the most preferred states according to $\mu_\mathcal{G}$. The pessimistic utility $E_*(d)$ is small as soon as there exists a possible consequence of $d$ which is both highly plausible and bad with respect to preferences. This is clearly a risk-averse and thus a pessimistic attitude. It has been shown in [7] that it is possible to compute $E_*(d)$ by only using a classical logic machinery on $\alpha$-level cuts of $\mathcal{K}_d$ and $\mathcal{G}$.

**Proposition 1** $E_*(d)$ *is the maximal value of $\alpha$ s.t.*

$$(\mathcal{K}_d)_\alpha \vdash (\mathcal{G})_{\underline{n(\alpha)}} \qquad (2)$$

*where $(B)_\alpha$, resp. $(B)_{\underline{\alpha}}$ is the set of classical propositions in a possibilistic logic base $B$ with a level greater or equal to $\alpha$, resp. strictly greater than $\alpha$.*

As seen in (2), $E_*(d)$ is equal to 1 ($\alpha = 1$) if the completely certain part of $\mathcal{K}_d$ entails the satisfaction of all the goals, even the ones with low priorities, since $\mathcal{G}_{\underline{0}}$ is just the set of all the propositions in $\mathcal{G}$ with a non-zero priority level. In [7] a computation procedure using an Assumption-based Truth Maintenance System is given for computing the best decision in the sense of (1)-(2).

An optimistic qualitative criterion (see Appendix) is given by

$$E^*(d) = \max_\omega \min(\mu_\mathcal{G}(\omega), \pi_{\mathcal{K}_d}(\omega)). \qquad (3)$$

The criterion $E^*(d)$ corresponds to an optimistic attitude since it is high as soon as there exists a possible consequence of $d$ which is both highly plausible and highly prized. $E^*(d)$ is equal to 1 as soon as one fully acceptable choice $\omega$ (i.e., such that $\mu_\mathcal{G}(\omega) = 1$) is also completely plausible. This criterion can also be expressed in logical terms.

**Proposition 2** $E^*(d)$ *is equal to the greatest $\alpha$ such that $(\mathcal{K}_d)_{\underline{n(\alpha)}}$ and $(\mathcal{G})_{\underline{n(\alpha)}}$ are logically consistent together.*

Let's consider the following example initially proposed by Boutillier in [5]. A similar example, but stated in a medical context, is used by Fox and Parsons [17].

**Example 1** *The example is about taking an umbrella or not, knowing that the sky is cloudy. The knowledge base is $\mathcal{K} = \{(u \rightarrow l, 1), (\neg u \rightarrow \neg l, 1), (u \rightarrow \neg w, 1), (r \wedge \neg u \rightarrow w, 1), (c, 1), (\neg r \rightarrow \neg w, 1), (c \rightarrow r, \lambda)\}$ ($0 < \lambda < 1$) with: l: to be overloaded, r: it rains, w: being wet, u: taking an umbrella, c: the sky is cloudy. The goals base is $\mathcal{G} = \{(\neg w, \overline{1}), (\neg l, \sigma)\}$ with ($\overline{0} < \sigma < \overline{1}$). We do not like to be overloaded with an umbrella, but it is more important to be dry. The set of decisions is $\mathcal{D} = \{u, \neg u\}$, i.e., taking an umbrella or not. The best pessimistic decision is to take an umbrella with $E_*(u) = n(\sigma)$. Moreover, $E_*(\neg u) = 0$, $E^*(u) = n(\sigma)$ and $E^*(\neg u) = n(\lambda)$. Thus the best decision in the optimistic case depends on the values $\lambda$ and $\sigma$.*

## 3 Argumentation-based decision making: Case of consistent bases

In this section, we suppose that the bases $\mathcal{K}$ and $\mathcal{G}$ are *consistent*. Consequently, the arguments will not be conflicting and thus the argumentation process is reduced to two steps: constructing arguments and concluding. Moreover, due to the use of two 'different bases', the arguments are defined in an 'original way'.

### 3.1 Pessimistic Criterion

In the pessimistic view, as pointed out by Proposition 1, we are interested in finding a decision $d$ (if it exists) such that $\mathcal{K}_\alpha \wedge d \vdash \mathcal{G}_\beta$ with $\alpha$ high and $\beta$ low, i.e. such that the decision $d$ together with the most certain part of $\mathcal{K}$ entails the satisfaction of the goals, even those with low priority. In this case, an argument supporting a decision takes the form of an *explanation*. The idea is that a decision is justified if it leads to the satisfaction of the most important goals, taking into account the most certain part of knowledge.

**Definition 1 (Argument PRO)** *An* argument *in favor of a decision d is a triple $A = <S, C, d>$ such that:*

- $d \in \mathcal{D}$
- $S \subseteq \mathcal{K}^*$ and $C \subseteq \mathcal{G}^*$
- $S \cup \{d\}$ *is consistent*
- $S \cup \{d\} \vdash C$
- $S$ *is minimal and $C$ is maximal (for set inclusion) among the sets satisfying the above conditions.*

$S = Support(A)$ is the support *of the argument, $C = Consequences(A)$ its* consequences *(the goals which are reached by the decision d) and $d = Conclusion(A)$ is the* conclusion *of the argument. The set $\mathcal{A}_P$ gathers all the arguments which can be constructed from $<\mathcal{K}, \mathcal{G}, \mathcal{D}>$.*

**Example 2 (Cont.)** *In the above example, there is one argument in favor of the decision 'u': $<\{u \rightarrow \neg w\}, \{\neg w\}, u>$. There is also a unique argument in favor of the decision '$\neg u$': $<\{\neg u \rightarrow \neg l\}, \{\neg l\}, \neg u>$.*

In [1, 20], it has been argued that arguments may have different strengths depending on the knowledge used to construct them. For instance, an argument built only from knowledge in $\mathcal{K}_1$ is stronger than an argument built using formula belonging to lower level cut of $\mathcal{K}$. In what follows, we will define the strengths of arguments in favor of decisions. In fact, an argument is evaluated from two points of view: according to the quality of knowledge used in the argument (the *certainty level* of the argument, for short the "*level*" in the following), and according to the goals satisfied by that argument (that is why we speak about the *degree of satisfaction* of the argument, for short the "*weight*" in the following). The level of an argument is the certainty degree of the less certain piece of knowledge used in that argument. The weight of an argument is all the greater as the priority degree of the most important goal which is not satisfied by the decision is small. Note that all the goals with a priority degree higher or equal to that degree should be satisfied by the decision. To compute the weight, we check the priority degree of the most important goal violated by the decision supported by that argument and we consider the degree which is immediately higher than it in the scale.

**Definition 2 (Strength of an Argument PRO)** *Let $A = <S, C, d>$ be an argument in $\mathcal{A}_P$. The strength of A is a pair $<Level_P(A), Weight_P(A)>$ such that:*

- *The* certainty level *of the argument is $Level_P(A) = min\{\rho_i \mid k_i \in S$ and $(k_i, \rho_i) \in \mathcal{K}\}$. If $S = \emptyset$ then $Level_P(A) = 1$.*

- *The* degree of satisfaction *of the argument is $Weight_P(A) = n(\beta)$ with $\beta = max\{\lambda_j \mid (g_j, \lambda_i) \in \mathcal{G}$ and $g_j \notin C\}$. If $\beta = \overline{1}$ then $Weight_P(A) = \overline{0}$ and if $C = \mathcal{G}^*$ then $Weight_P(A) = \overline{1}$.*

**Example 3 (Cont.)** *In the above example, the level of the argument $<\{u \rightarrow \neg w\}, \{\neg w\}, u>$ is 1 whereas its weight is $n(\sigma)$. Concerning the argument $<\{\neg u \rightarrow \neg l\}, \{\neg l\}, \neg u>$, its level is 1 and its weight is $n(\overline{1}) = \overline{0}$.*

The strengths of arguments make it possible to compare pairs of arguments as follows:

**Definition 3** *Let A and B be two arguments in $\mathcal{A}_P$. A is preferred to B iff $min(Level_P(A), Weight_P(A)) \geq min(Level_P(B), Weight_P(B))$.*

**Example 4 (Cont.)** *In our example, the argument $<\{u \to \neg w\}, \{\neg w\}, u>$ is preferred to the argument $<\{\neg u \to \neg l\}, \{\neg l\}, \neg u>$.*

As shown before, arguments are constructed in favor of decisions and those arguments can be compared. In what follows, we show that decisions can also be compared on the basis of the relevant arguments.

**Definition 4** *Let $d, d' \in \mathcal{D}$. $d$ is* preferred *to $d'$ iff $\exists A \in \mathcal{A}_P$, $Conclusion(A) = d$ such that $\forall B \in \mathcal{A}_P$, $Conclusion(B) = d'$, then $A$ is preferred to $B$.*

**Example 5 (Cont.)** *In our example, the decision $u$ is preferred to the decision $\neg u$ since $<\{u \to \neg w\}, \{\neg w\}, u>$ is preferred to $<\{\neg u \to \neg l\}, \{\neg l\}, \neg u>$.*

The following result relates the strengths of arguments and the preference relations between them to the pessimistic qualitative utility.

**Theorem 1** *Let $d \in \mathcal{D}$. $E_*(d) \geq \alpha$ iff $\exists A \in \mathcal{A}_P$ such that $Conclusion(A) = d$ and $min(Level_P(A), Weight_P(A)) = \alpha$.*

**Corollary 1** *Let $d \in \mathcal{D}$. $E_*(d) = \alpha$ iff $\exists A \in \mathcal{A}_P$ with $Conclusion(A) = d$ and $min(Level_P(A), Weight_P(A)) = \alpha$ such that $A$ is preferred to any argument $A'$ with $Conclusion(A') = d$.*

### 3.2 Optimistic Criterion

In the optimistic point of view, we are interested in finding a decision $d$ (if it exists) which is consistent with the knowledge base and the goals (i.e. $\mathcal{K}^* \wedge \{d\} \wedge \mathcal{G}^* \neq \bot$). This is optimistic in the sense that it assumes that goals may be attained as soon as their negation cannot be proved.

Unlike the pessimistic case, in the optimistic point of view we will consider the arguments which show clearly that goals will not be satisfied by the decision. Indeed, a decision will be considered as being all the better as there does not exist any strong argument against it. An argument against a decision is defined as follows:

**Definition 5 (Argument CON)** *An* argument *against a decision $d$ is a triple $A = <S, C, d>$ such that:*

- $d \in \mathcal{D}$
- $S \subseteq \mathcal{K}^*$ *and* $C \subseteq \mathcal{G}^*$
- $S \cup \{d\}$ *is consistent*
- $\forall g_i \in C, S \cup \{d\} \vdash \neg g_i$
- $S$ *is minimal and $C$ is maximal (for set inclusion) among the sets satisfying the above conditions.*

$S = Support(A)$ is the support *of the argument, $C = Consequences(A)$ its* consequences *(the goals which are not satisfied by the decision $d$), and $d = Conclusion(A)$ its conclusion. The set $\mathcal{A}_O$ gathers all the arguments which can be constructed from $<\mathcal{K}, \mathcal{G}, \mathcal{D}>$.*

Note that the consequences considered here are the negative ones.

**Example 6 (Cont.)** *In the above example, there is one argument against the decision 'u': $<\{u \to l\}, \{\neg l\}, u>$. There is also a unique argument against the decision $\neg u$: $<\{c, c \to r, r \wedge \neg u \to w\}, \{\neg w\}, \neg u>$.*

How do we compute the levels and weights of such arguments? An argument against a decision is all the stronger that it is based on the most certain part of the knowledge and that it attacks a more important goal. An argument against "'d'" will be all the weaker as it requires the use of weak knowledge or if it only attacks low priority goals. This leads us to define the weakness of an argument CON.

**Definition 6 (Weakness of an Argument CON)** *Let $A = <S, C, d>$ be an argument of $\mathcal{A}_O$. The weakness of $A$ is a pair $<Level_O(A), Weight_O(A)>$ such that:*

- *The* level *of the argument is $Level_O(A) = n(\varphi)$ such that $\varphi = min\{\rho_i \mid k_i \in S$ and $(k_i, \rho_i) \in \mathcal{K}\}$. If $S = \emptyset$ then $Level_O(A) = 0$.*

- *The* degree *of the argument is $Weight_O(A) = n(\beta)$ such that $\beta = max\{\lambda_j$ such that $g_j \in C$ and $(g_j, \lambda_i) \in \mathcal{G}\}$.*

**Example 7 (Cont.)** *In the above example, the level of the argument $<\{u \to l\}, \{\neg l\}, u>$ is 0 whereas its degree is $n(\sigma)$. Concerning the argument $<\{c, c \to r, r \wedge \neg u \to w\}, \{\neg w\}, \neg u>$, its level is $n(\lambda)$, and its degree is 0.*

Once we have defined the arguments and their weaknesses, we are now ready to compare pairs of arguments. Since we are interested in decisions for which all the arguments against it are weak, we are interested in the least weak arguments against a considered decision. This leads to the two following definitions:

**Definition 7** *Let $A$ and $B$ be two arguments in $\mathcal{A}_O$. $A$ is* preferred *to $B$ iff $max(Level_O(A), Weight_O(A)) \geq max(Level_O(B), Weight_O(B))$.*

**Example 8 (Cont.)** *In the above example, the comparison of the two arguments amounts to compare $n(\sigma)$ with $n(\lambda)$.*

As in the pessimistic case, decisions will be compared on the basis of the relevant arguments.

**Definition 8** *Let $d, d' \in \mathcal{D}$. $d$ is* preferred *to $d'$ iff $\exists A \in \mathcal{A}_O$ with $Conclusion(A) = d$ such that $\forall B \in \mathcal{A}_O$ with $Conclusion(B) = d'$, then $A$ is preferred to $B$.*

**Example 9 (Cont.)** *In the above example, the comparison of the two decisions $u$ and $\neg u$ depends on the respective values of $\sigma$ and $\lambda$. Namely, if $\sigma$ (the priority of the goal "not overloaded") is small then the best decision will be to take an umbrella. If certainty degree $\lambda$ of having rain is small enough then the best optimistic decision will not be to take an umbrella.*

The following result relates the weaknesses of arguments and the preference relations between them to the optimistic qualitative utility.

**Theorem 2** *Let $d \in \mathcal{D}$. $E^*(d) \geq \alpha$ iff $\exists A \in \mathcal{A}_O$ with $Conclusion(A) = d$ such that $max(Level_O(A), Weight_O(A)) = \alpha$.*

**Corollary 2** *Let $d \in \mathcal{D}$. $E^*(d) = \alpha$ iff $\exists A \in \mathcal{A}_O$ with $Conclusion(A) = d$ and $max(Level_O(A), Weight_O(A)) = \alpha$ such that $A$ is preferred to any argument $A'$ with $Conclusion(A') = d$.*

## 4 Argumentation-based decision making: Case of inconsistent bases

In section 3, we have assumed the consistency of both $\mathcal{K}$ and $\mathcal{G}$ which is required for Theorem 1 and Theorem 2. However, the described argumentative approach still makes sense when $\mathcal{G}$ is inconsistent. When $\mathcal{K}$ is also inconsistent things seem different since two problems should be handled at the same time: the decision problem and the one of handling inconsistency in $\mathcal{K}$. In [1], an argumentation framework has been proposed for handling inconsistency in a knowledge base. In this section, we propose a new framework which computes the 'best' decision (if it exists) by combining ideas from [1] and from the above section. In the case of pessimistic criterion, two kinds of arguments can be defined:

- Arguments in favor of decisions (see definition 1). The set $\mathcal{A}_p$ gathers all those arguments.
- Arguments in favor of pieces of knowledge (or beliefs). In classical argumentation frameworks for handling inconsistency in knowledge bases, such arguments are seen as logical proofs of the beliefs they support.

**Definition 9 (Argument in favor of a belief)** *An argument in favor of a belief is a pair $A = <H, h>$ such that:*

- $H \subseteq \mathcal{K}^*$
- $H \vdash h$
- *$H$ is consistent and minimal (for set inclusion) among the sets satisfying the above conditions.*

*$H = Support(A)$ is the support of the argument and $h = Conclusion(A)$ its conclusion. The $\mathcal{A}$ gathers all the arguments which can be constructed from $\mathcal{K}$.*

As mentioned before, the base $\mathcal{K}$ is pervaded with uncertainty. From the certainty degrees, we define the certainty level of an argument in favor of a belief.

**Definition 10 (Certainty level)** *Let $A = <H, h> \in \mathcal{A}$. The certainty level of $A$, denoted by $Level(A) = min\{\rho_i \mid k_i \in H$ and $(k_i, \rho_i) \in \mathcal{K}\}$. If $H = \emptyset$ then $Level(A) = 1$.*

In [1], arguments are compared according to their certainty levels. Hence, some arguments are preferred to others.

**Definition 11** *Let $A$ and $B \in \mathcal{A}$. $A$ is preferred to $B$ iff $Level(A) \geq Level(B)$.*

An argument in favor of a belief can also be compared with an argument in favor of a pessimistic decision as follows:

**Definition 12** *Let $A \in \mathcal{A}$ and $B \in \mathcal{A}_P$. $A$ is preferred to $B$ iff $Level(A) \geq Level_P(B)$.*

In general, since $\mathcal{K}$ is inconsistent, arguments in $\mathcal{A}$ will conflict. Moreover, these arguments may also conflict with arguments in favor of decisions. We make this idea precise with the notions of *undercut* and *attack*:

**Definition 13** *Let $A_1$ and $A_2 \in \mathcal{A}$ and $A_3 \in \mathcal{A}_P$.*

- *$A_1$ undercuts $A_2$ iff $\exists h \in Support(A_2)$ such that $Conclusion(A_1) \equiv \neg h$. In other words, an argument is undercut iff there exists an argument for the negation of an element of its support.*

- *$A_1$ attacks $A_3$ iff $\exists h \in Support(A_3)$ or $\exists h \in Consequences(A_3)$ such that $Conclusion(A_1) \equiv \neg h$.*

We can now define the argumentation system we will use:

**Definition 14 (Argumentation system)** *An argumentation system is a tuple $<\mathcal{A}_P, \mathcal{A}, Undercut, Attack>$. This system gives raise to three classes of arguments:*

- *The class $\underline{S}$ of acceptable arguments.*

- *The class $\underline{R}$ of rejected arguments. Such arguments are undercut or attacked by acceptable ones.*

- *The class $\underline{C}$ of arguments in abeyance. Such arguments are neither acceptable nor rejected. $\underline{C} = (\mathcal{A} \cup \mathcal{A}_P) \setminus (\underline{S} \cup \underline{R})$.*

In what follows, we will start by defining the acceptable arguments. The levels of the arguments make it possible to distinguish different types of relations between arguments (in favor of beliefs or decisions):

**Definition 15** *Let $A$, $B \in \mathcal{A} \cup \mathcal{A}_P$, and $\mathcal{V} \subseteq \mathcal{A} \cup \mathcal{A}_P$.*

- *$B$ strongly undercuts $A$ (resp. $B$ strongly attacks $A$) iff $B$ undercuts $A$ (resp. $B$ attacks $A$) and it is not the case that $A$ is preferred to $B$.*

- *If $B$ undercuts $A$ (resp. $B$ attacks $A$) then $A$ defends itself against $B$ iff $A$ is preferred to $B$.*

- *A set of arguments $\mathcal{V}$ defends an argument $A$ if there is some argument in $\mathcal{V}$ which strongly undercuts (resp. strongly attacks) every argument $B$ where $B$ undercuts (resp. attacks) $A$ and $A$ cannot defend itself against $B$.*

Let $C$ gather all non-undercut and non-attacked arguments and arguments defending themselves against all their undercutting arguments and against all their attackers. In [2], it was shown that the set of acceptable arguments in favor of beliefs is the least fixpoint of a function $\mathcal{F}$:

$$\begin{aligned} \mathcal{S} &\subseteq \mathcal{A}, \\ \mathcal{F}(\mathcal{S}) &= \{(H, h) \in \mathcal{A} | (H, h) \text{ is defended by } \mathcal{S}\}\end{aligned}$$

This definition is generalized to the case of handling two types of arguments. Formally:

**Definition 16** *The set of* acceptable *arguments in favor of beliefs and in favor of decisions is:*

$$\underline{S} = \bigcup \mathcal{F}_{i \geq 0}(\emptyset) = C \cup \left[ \bigcup \mathcal{F}_{i \geq 1}(C) \right]$$

Once we have defined the different categories of arguments, we can now define the different categories of decisions.

**Definition 17** *Let $d \in \mathcal{D}$.*

- *$d$ is a* rejected *decision iff $\forall\ A \in \mathcal{A}_P$ such that $Conclusion(A) = d$ then $A \in \underline{R}$.*

- *$d$ is a* candidate *decision iff $\exists\ A \in \underline{S}$ such that $Conclusion(A) = d$.*

The candidate decisions can be compared as follows:

**Definition 18** *Let $d$ and $d' \in \mathcal{D}$. $d$ is preferred to $d'$ iff $\exists\ A \in \underline{S}$ and $Conclusion(A) = d$ such that $\forall\ A' \in \underline{S}$ and $Conclusion(A') = d'$, then $min(Level_P(A), Weight_P(A)) \geq min(Level_P(A'), Weight_P(A'))$.*

Note that when $\mathcal{K}$ is consistent, we retrieve the decision procedure described in Section 3.

## 5　Related works

As said in the introduction, some works have been done on arguing for decision. In [17], no explicit distinction is made between knowledge and goals. However, in their examples, values (belonging to a linearly ordered scale) are assigned to formulas which represent goals. These values provide an empirical basis for comparing arguments using a symbolic combination of strengths of beliefs and goals values. This symbolic combination is performed through dictionaries corresponding to different kinds of scales that may be used.

In [4], Bonet and Geffner have also proposed an original approach to qualitative decision, inspired from Tan and Pearl [21], based on "action rules" that link a situation and an action with the satisfaction of a *positive* or a *negative* goal. However in contrast with the previous work and the work presented in this paper, this approach does not refer to any model in argumentative inference.

In their framework, there are four parts:

1. a set $\mathcal{D}$ of *actions* or *decisions*.

2. a set $\mathcal{I}$ of input propositions defining the possible input situation. A degree of plausibility is associated with each input. Thus, $\mathcal{I} = \{(k_i, \alpha_i)\}$ with $\alpha_i \in \{likely, plausible, unlikely\}$.

3. a set $\mathcal{G}$ of *prioritized goals* such that $\mathcal{G} = \mathcal{G}^+ \cup \mathcal{G}^-$. $\mathcal{G}^+$ gathers the *positive* goals that one wants to achieve and $\mathcal{G}^-$ gathers the *negative* goals that one wants to avoid. Thus, $\mathcal{G} = \{(g_i, \beta_i)\}$ with $\beta_i \in [0, 1, ...N]$.
Note that in our framework what they call here negative goals are considered in our goal base as negative literals.

4. a set of *action rules* $\mathcal{AR} = \{(A_i \wedge C_i \Rightarrow x_i, \lambda_i), \lambda_i \geq 0\}$, where $A_i$ is an action, $C_i$ is a conjunction of input literals, and $x_i$ is a goal. Each action rule has two measures: a *priority* degree which is exactly the priority degree of the goal $x_i$, and a *plausibility* degree. This plausibility is defined as follows:

    - a rule $A \wedge C \Rightarrow x$ is likely if any conjunct of $C$ is likely.
    - a rule $A \wedge C \Rightarrow x$ is unlikely if some conjunct of $C$ is unlikely.
    - a rule $A \wedge C \Rightarrow x$ is plausible if it is neither likely nor unlikely.

In this approach only input propositions are weighted in terms of plausibility. Action rules inherit these weights through the three above rules in a rather empirical manner which depends on the chosen plausibility scale. The action rules themselves are not weighted since they are potentially understood as defeasible rules, although no non-monotonic reasoning system is associated with them. Contrastedly, in our approach, we use an abstract scale. Moreover, weighted possibilistic clauses have been shown to be able to properly handle non-monotonic inference in the sense of Kraus, Lehmann and Magidor [19]' preferential system augmented with rational monotony (see [3]). So a part of our weighted knowledge may be viewed as the encoding of a set of default rules. From the above four bases, reasons are constructed for (against) actions in [4]. Indeed, goals provide reasons for (or against) actions. Positive goals provide reasons *for* actions, whereas negative goals provide reasons *against* actions. The basic idea behind this distinction is that negative goals should be discarded, and consequently any action which may lead to the satisfaction of such goals should be avoided. However, the approach makes no distinction between what we call pessimism and optimism. The definition of a "'reason'" in [4] is quite different from our definition of an argument. Firstly, a reason considers only one goal and secondly, the definition is poor since it only involves facts. Finally, in Bonet and Geffner's framework, decisions which satisfy the most important goals are privileged. This is also true in our approach, but the comparison between decisions can be further refined, in case of several decisions yielding to the satisfaction of the most important goals, by taking into account the other goals which are not violated by these decisions.

## 6 Conclusion

The paper has sketched a method, agreeing with qualitative possibility-based decision, which enables us to compute and justify best decision choices. We have shown that it is possible to design a logical machinery which directly manipulates arguments with their strengths and compute acceptable and best decisions from them. The approach can be extended in different various directions. The computation of the strengths of arguments pro and con can be refined by using vector of values rather than scalar values for refining max and min aggregation [15], in order to take into account the presence of several weak points in an argument for instance. Another extension of this work consists of allowing for inconsistent knowledge or goal bases as preliminarily discussed in section 4. We are now working on this point. We are also planning to transpose the approach to multiple criteria decision making from the one proposed here for decision under uncertainty, taking advantage of the close relation between both areas [9].

## 7 Acknowledgments

This work was supported by the Commission of the European Communities under contract IST-2004-002307, ASPIC project "Argumentation Service Platform with Integrated Components".

## 8 Appendix: Pessimistic and optimistic decision criteria

Normalized possibility distributions which map a set of interpretations to a scale are a convenient way of encoding complete pre-orderings. It makes sense, if information is qualitative, to represent incomplete knowledge on the actual state by a possibility distribution $\pi$ on $S$, the set of (mutually exclusive) states, with values in a plausibility scale $R$, and the decision-maker's preferences on the set $X$ of (mutually exclusive) consequences, by means of another possibility distribution $\mu$ with values on a preference scale $T$. Let 0 (resp. $\overline{0}$) and 1 (resp. $\overline{1}$) denote the bottom and top elements of $R$ (resp. $T$). The following representational conventions are assumed for possibility distributions. $\pi(s) = 0$ means that $s$ is definitely impossible according to what is known, and $\mu(x) = \overline{0}$ that $x$ is unacceptable as a consequence. The greater $\pi(s)$ (resp. $\mu(x)$), the more plausible $s$ as being the real state of the world (resp. the more acceptable $x$ as a consequence). $\pi(s) = 1$ (resp. $\mu(x) = \overline{1}$) means $s$ is among the most plausible (normal) states and there may be several $s$ such that $\pi(s) = 1$ (resp. $x$ is among the most preferred consequences). The utility of a decision $f$ whose consequence in state $s$ is $x = f(s) \in X$, for all states $s$, can be evaluated by combining the plausibilities $\pi(s)$ and the utilities $\mu(x)$ in a suitable way. Two qualitative criteria that evaluate the worth of decision $f$ have been proposed in the literature, provided that a commensurateness assumption between plausibility and preference is made:

**Definition 19** *Pessimistic criterion*

$$E_*(f) = inf_{s \in S} \max(n(\pi(s)), \mu(f(s))), \qquad (4)$$

where $n$ is an order-reversing mapping from $R$ to $T$ (i.e. $n(0) = \overline{1}$, $n(1) = \overline{0}$, $n$ is a strictly decreasing bijection of $R$ to $T$).

**Definition 20** *Optimistic criterion*

$$E^*(f) = sup_{s \in S} \min(m(\pi(s)), \mu(f(s))), \qquad (5)$$

where $m$ is an order-preserving map from $R$ to $T$ (i.e. $m(0) = \overline{0}$, $m(1) = \overline{1}$ and $m$ is strictly increasing).

These criteria are nothing but the necessity and the possibility measures of fuzzy events, and are special cases of Sugeno integrals [11]. Maximizing $E_*(f)$ means finding a decision $f$, all the highly plausible consequences of which are also highly preferred. The definition of "highly plausible" is decision-dependent and reflects the compromise between high plausibility and low utility expressed by the order-reversing map between the plausibility valuation set $R$ and the utility valuation set $T$. It generalizes the maxmin criterion, which is based on the worst possible consequence of the considered decision in the absence of probabilistic knowledge, since if $\pi$ is then the set characteristic function of a subset $A$ of states, $E_*(f)$ is the utility of the worst consequence of states in $A$, however unlikely they are. But the possibilistic criterion is less pessimistic. It focuses on the idea of usuality since it relies on the worst *plausible* consequences induced by the decision (extremely unusual consequences are neglected). $E^*(f)$ generalizes the maximax optimistic criterion. The latter evaluation can be used as a secondary criterion, for breaking ties between decisions which are equivalent w.r.t. the pessimistic criterion.